\documentclass[lettersize,journal]{IEEEtran}
\usepackage[authoryear,longnamesfirst]{natbib}
\usepackage{xcolor}
\usepackage{float}
\usepackage{pifont}
\usepackage{booktabs} 
\usepackage{multirow}
\usepackage{xspace}
\usepackage{enumitem}
\usepackage{balance}
\usepackage{caption}
\usepackage{listings}
\usepackage{tikz}
\usepackage{amsmath,amssymb,amsfonts}
\usepackage{algorithmic}
\usepackage{filecontents}

\usepackage{pifont}
\usepackage{graphicx}
\usepackage{textcomp}
\usepackage{adjustbox}
\usepackage{lscape}
\usepackage{tabularx}
\usepackage{color}
\usepackage[newcommands]{ragged2e}
\usepackage{colortbl}
\usepackage{array}
\usepackage{float}
\usepackage{url} 
\usepackage{subcaption,lipsum,booktabs}
\usepackage{siunitx} 

\usepackage{kantlipsum} 
\usepackage{mwe} 

\usepackage{algorithm}
\usepackage{algorithmic}

\def\tsc#1{\csdef{#1}{\textsc{\lowercase{#1}}\xspace}}
\tsc{WGM}
\tsc{QE}
\tsc{EP}
\tsc{PMS}
\tsc{BEC}
\tsc{DE}

\title{FedRDF: A Robust and Dynamic Aggregation Function against Poisoning Attacks in Federated Learning}

\author{Enrique Mármol Campos, Aurora Gonzalez-Vidal Jos\'{e} L. Hern\'{a}ndez-Ramos and Antonio Skarmeta

\thanks{Enrique Marmol, Aurora Gonzalez-Vidal Jos\'{e} L. Hern\'{a}ndez-Ramos and Antonio Skarmeta are with the University of Murcia, Spain. E-mail: \{enrique.marmol, aurora.gonzalez2, jluis.hernandez, skarmeta\}@um.es}}

\begin{document}
\let\WriteBookmarks\relax
\def\floatpagepagefraction{1}
\def\textpagefraction{.001}
\newcommand{\PreserveBackslash}[1]{\let\temp=\\#1\let\\=\temp}
\newcolumntype{C}[1]{>{\PreserveBackslash\centering}p{#1}}
\newcolumntype{R}[1]{>{\PreserveBackslash\raggedleft}p{#1}}
\newcolumntype{L}[1]{>{\PreserveBackslash\raggedright}p{#1}}

\markboth{IEEE}{FedRDF: A Robust and Dynamic Aggregation Function against Poisoning Attacks in Federated Learning}
\maketitle

\begin{abstract}
Federated Learning (FL) represents a promising approach to typical privacy concerns associated with centralized Machine Learning (ML) deployments. Despite its well-known advantages, FL is vulnerable to security attacks such as Byzantine behaviors and poisoning attacks, which can significantly degrade model performance and hinder convergence. The effectiveness of existing approaches to mitigate complex attacks, such as median, trimmed mean, or Krum aggregation functions, has been only partially demonstrated in the case of specific attacks. Our study introduces a novel robust aggregation mechanism utilizing the Fourier Transform (FT), which is able to effectively handling sophisticated attacks without prior knowledge of the number of attackers. Employing this data technique, weights generated by FL clients are projected into the frequency domain to ascertain their density function, selecting the one exhibiting the highest frequency. Consequently, malicious clients' weights are excluded. Our proposed approach was tested against various model poisoning attacks, demonstrating superior performance over state-of-the-art aggregation methods. 

\end{abstract}


%


\IEEEpeerreviewmaketitle

\maketitle

\section{Introduction}

\IEEEPARstart{F}{ederated} Learning (FL) \cite{mcmahan2017communication} is a decentralized paradigm of Machine Learning (ML) where data is not shared through typical datacenters for further analysis. FL involves multiple clients (or parties) and an aggregator or set of aggregators. Such clients collaboratively train an ML model using their respective datasets without sharing the raw data; indeed, solely the parameters generated during local training are shared with the aggregator. This decentralized ML approach addresses some of the main privacy challenges associated with conventional centralized ML deployments in terms of privacy and its impact on the enforcement of General Data Protection Regulation (GDPR) principles \cite{goddard2017eu}. In this context, FL emerges as a promising solution for privacy-related concerns and also for mitigating the issues associated with the delay required to share and process the data on servers or centralized platforms 
during model training \cite{rasha2023federated}. Consequently, FL has found widespread utility across multiple domains \cite{rodriguez2023survey}, including finance, healthcare, smart cities \cite{matheu2022federated} and cybersecurity \cite{campos2021evaluating}.

%
The training process in a FL scenario encompasses each client transmitting their individual model updates or weights to the aggregator, which subsequently aggregates these contributions by using a certain \textit{aggregation function}.Typically, the aggregation function entails averaging the clients' weights and sending the outcome back to the clients in each training round. However, like in centralized ML deployments, FL is subject to various types of security attacks, including Byzantine clients, which do not behave as expected. Specifically, \textit{poisoning attacks} are intended to modify the data or directly the weights shared by clients with the aggregator to degrade the model's performance and hinder its convergence \cite{xia2023poisoning}. Indeed, previous works confirm FL's susceptibility to poisoning attacks \cite{baruch2019little}, particularly highlighting the vulnerability of the average function or FedAvg \cite{mcmahan2017communication} as an aggregation approach. In the context of FL, model poisoning attacks involve maliciously modifying the weights/gradients in each training round. As we discussed in \cite{hernandez2023intrusion}, these attacks are especially relevant in FL. In fact, since the data is not shared, model poisoning also includes the effect produced by data poisoning attacks, such as dirty-label \cite{bhagoji2019analyzing}. To address these attacks, various robust aggregation functions were proposed, such as the use of median, trimmed mean \cite{yin2018byzantine} or Krum \cite{blanchard2017machine}. However, most of these approaches, as well as the works based on these techniques, only address simple poisoning attacks, where attackers lack the ability to collude with each other. Additionally, some of these proposals assume that the number of attackers is known, which may not be feasible in real-world scenarios or against sophisticated attacks where the compromised nodes can change in each training round.

Unlike the previous approaches, our robust aggregation mechanism does not require knowledge of the number of attackers in a certain scenario, and it is not based on distance or statistical approaches. Instead, we employ a technique based on the well-known Fourier Transform (FT) \cite{gray2012fourier}, which is used to convert the weights sent in each round into the frequency domain to calculate their density function. Thus, only the weights with higher frequencies are selected. Our approach has been thoroughly evaluated against two types of model poisoning attacks: in the first case, compromised clients generate random weights to be shared with the aggregator; in the second case, we use the min-max attack where the compromised nodes collude to achieve a greater impact on the performance of the resulting model \cite{shejwalkar2021manipulating}. The evaluation results demonstrate that our approach offers better results against both attacks compared to well-known aggregation functions, without adding further complexity. Additionally, our proposed FT-based approach is complemented with a simple mechanism to determine the existence of Byzantine attackers in each training round. The main reason is that in the absence of attackers, the use of FedAvg as an aggregation mechanism provides better results compared to robust aggregation functions. Thus, the resulting approach FedRDF is designed to maximize the effectiveness of the system by choosing between FedAvg and FT depending on the presence of malicious clients in each training round. In summary, the contributions of our work are:
\begin{itemize}
    \item Design and implementation of a novel Fourier Transform-based robust aggregation function for model poisoning attacks

    \item Definition of a threat model and implementation of different attacks for the evaluation and analysis of the proposed approach 
    
    \item Dynamic aggregation strategy to alternate between FedAvg and the proposed FT-based method based on the detection of malicious clients

   \item Extensive performance evaluation demonstrating enhanced resistance against certain attacks compared with state-of-the-art approaches
\end{itemize}

The paper's organization is as follows: it starts with foundational concepts in Section \ref{sec:Preliminaries}, then explores related works in Section \ref{related} focusing on robust aggregation against attacks. Section \ref{sec:Setup} outlines the threat model and considered attacks. Following this, Section \ref{sec:methodology} details the robust aggregation steps and dynamic client selection. The evaluation results are presented in Section \ref{sec:evaluation}, and finally, Section \ref{sec:conclution} summarizes the study's findings and conclusions.


\section{Preliminaries}\label{sec:Preliminaries}
In this section, we describe the main notions related to our work, including the FL components and process, as well as the role of aggregation functions and the typical structure of poisoning attacks.

\subsection{Federated learning}
Federated learning (FL) \cite{mcmahan2017communication} introduced a decentralized collaborative approach in ML, enabling model training without sharing underlying data. FL settings typically comprise distinct entities: the \textit{clients} and an \textit{aggregator}. Each client possesses unique, private datasets undisclosed to others during the collaborative training process. Clients collectively train a shared ML model, solely exchanging the resultant weights while the aggregator, or server, aggregates these weights and redistributes them for subsequent training. This FL process unfolds through distinct steps: 

\begin{enumerate}
    \item The aggregator initializes the model and dispatches it to the clients.
    \item Clients individually train the model through local iterations (epochs).
    \item Clients transmit their obtained weights to the aggregator.
    \item The aggregator aggregates these weights to generate new weights, subsequently relayed to the clients.
\end{enumerate}

Steps 2), 3), and 4) are iterated multiple times (rounds) until weight convergence is achieved.


The pivotal element within the server is the \textit{aggregation function}, commonly exemplified by FedAvg, initially proposed by \cite{mcmahan2017communication}, in which weights are aggregated trough the mean. FedAvg involves clients calculating local weights and transmitting them to the server. The server aggregates these weights by computing their average based on the respective data lengths. Specifically, the FedAvg objective function is derived using Equation \ref{eq:FedAvg}, wherein $W = (w_i)_{i=1}^n$ denotes the server's model weights, $W^k= (w_i^k)_{i=1}^n$ represents client $k$'s weights, $D_k$ is the length of the client $k$ dataset, and $D$ total sum of all dataset lengths. 

\begin{equation} \label{eq:FedAvg}
    W =\sum_{k=1}^{K} \frac{D_k}{D} W^k,
\end{equation}


\subsection{Poisoning attacks}
Poisoning attacks in FL are intended to harm the training phase by introducing maliciously modified data or weights to undermine the model performance \cite{barreno2006can}. These attacks are usually represented in two forms: data poisoning attacks and local model poisoning attacks. Data poisoning involves manipulating a client's local data to induce misclassification. Conversely, local model poisoning targets altering the weights generated through the local training process, so that clients transmit corrupted weights to the aggregator. Recent works \cite{bhagoji2019analyzing, fang2018poisoning} argue that data poisoning attacks can essentially be perceived as local model poisoning by computing their corresponding weight modifications. Moreover, empirical evidence suggests that local model poisoning attacks yield more significant impact than dataset alterations. Consequently, this study focuses on attacks intended to manipulate clients' weights.

Local model poisoning attacks encompass two classes: untargeted and targeted poisoning attacks \cite{tian2022comprehensive}. Untargeted attacks constitute the basic form, seeking to minimally modify the global model to yield incorrect predictions for different samples. In contrast, targeted attacks are intricate, introducing a multi-task problem to manipulate the victim's predictions on specific samples while maintaining functionality on benign samples—distinct from random alterations in untargeted attacks. This study prioritizes untargeted attacks to safeguard the system against any form of assault, unlike targeting specific misclassifications. Additionally, research in \cite{kiourti2020trojdrl} underscores the challenge in detecting untargeted poisoning attacks compared to their targeted counterparts. Section \ref{sec:methodology} provides comprehensive insight into the attack types considered within this study.

\subsection{Robust aggregations functions}
Robust aggregation functions are meant to resist corrupted weights by looking for alternatives to the mean, aggregation functions that can filter the correct weights from the false ones. We show a summary of the robust aggregation functions considered in this work in table \ref{tab:aggFun}. In this table, the main characteristics of these functions are displayed, including the underlying technique, the requirement of known number of attackers, and the computational cost.

\begin{table*}[]
\centering
\begin{tabular}{ccccc}
 & \textbf{Underlying technique}  & \textbf{Requirement of known number of attackers} & \textbf{Computational cost} \\ \hline
\textbf{Median}       & Median statistic    & No  & Medium \\ \hline
\textbf{Krumm}        & Euclidean distance      & Yes & High   \\ \hline
\textbf{Trimmed mean} & Validation data       & Yes & Medium \\ \hline
\textbf{FFT}          & Frequency statistics   & No  & Medium \\ \hline
\end{tabular}
\caption{Summary of the main state-of-the-art aggregation functions and our FFT-based approach}
\label{tab:aggFun}
\end{table*}

\subsubsection{FedMedian} 
FedMedian, proposed by \cite{yin2018byzantine}, represents a different approach compared with FedAvg by employing a median computation rather than averaging weights. This variant computes the median in a coordinate-wise manner, evaluating weight values individually.

\begin{equation}\label{eq:median}
    W= Median(W^1,\dots,W^K)
\end{equation}

\subsubsection{Trimmed Mean} 

The trimmed mean, as proposed by \cite{yin2018byzantine}, constitutes a distinct approach to mean computation. This method involves selecting a value $n<\frac{K}{2}$, where $K$ represents the number of clients. Subsequently, in a coordinate-wise manner, the server eliminates the lowest $n$ values and the highest $n$ values, computing the mean from the remaining $K-2n$ values. The selection of $n$ aims to eliminate outlayers and enhance the system's capability against potential adversarial behavior from $n$ malicious clients, thereby fortifying the system against up to 50\% of malicious entities. Denoting this resulting set as $I$:

\begin{equation}\label{eq:tm}
    W = \sum_n^{K-n} \frac{|I_i|}{|I|} W^i,
\end{equation}

\subsubsection{Krumm}
The Krumm aggregation function, introduced by \cite{blanchard2017machine}, derives from the Krumm function (Equation \ref{eq:krum}). This function operates on the clients' weights $W^k$, selecting the weight with the lowest sum among their $K-f-2$ nearest neighbors. Here, $K$ denotes the total number of clients, $f$ represents a server-determined constant indicative of the count of malicious clients, and $\Gamma_{K-f-2}^i$ denotes the set of nearest neighbors for client $k$. Subsequently, the server transmits this selected value to the remaining clients.

\begin{equation}\label{eq:krum}
    Krum(W^1, W^2, \dots, W^K) = Min\{(\sum_{j\in \Gamma_{K-f-2}^i} ||W^i - W^j||)_{i=1}^{K}\}
\end{equation}

\subsection{Data representation}

There are several approaches to representing a numeric variable, as the set of weights in a FL environment are. The most basic one is to compute the mean and standard deviation among other
statistical measures (e.g. variance, mode) and that is how FedAvg, FedMean and other \textit{basic} functions are composed. However, using those statistics it is not possible to represent and grasp all the information that the variable contains \cite{gonzalez2018beats}.

Representation methods are techniques used to transform input data into a more meaningful and informative representation by means of symbols \cite{lin2007experiencing}, aggregate by intervals \cite{zhang2011piecewise}, domain transformations \cite{gonzalez2018beats}, and other kinds of operations.

If we interpret aggregation functions as a way to represent the set of weights, it is possible to use more elaborated representation methods in order to perform aggregation in FL scenarios. The only restriction is that those methods need to be invertible, meaning that they should allow for the reconstruction or decoding of the original input data from the learned representation so that the final aggregated result is in the domain of the original weights. This restriction needs to be fulfilled so that the aggregation can be used again in the different clients as weight without compromising the clients' models.  Some examples of invertible transformation are the Discrete Cosine Transform (DCT), Autoencoders, Principal Component Analysis (PCA) and the Fourier Transform.

\subsubsection{Fourier Transform}\label{sec:fourier}

The Fourier transform emerges as a specialized case within the realm of Fourier series \cite{tolstov2012fourier}. Fourier series find application in analyzing periodic functions over $\mathbb{R}$, specifically those conforming to $f(x)=f(z+L)$, defined by Equation \ref{eq:serieFourier}.

\begin{equation}\label{eq:serieFourier}
f(x) = \sum_{n=-\infty}^\infty \hat{f}(n)\exp^{\frac{2\pi i n x}{L}},
\end{equation}

Here, $\hat{f}(n) = \frac{1}{L}\int_a^bf(x)\exp^{\frac{-2\pi i n x}{L}} dx$ , $n \in \mathbb{Z}$ , and  $L=b-a$.

The Fourier transform extends this theory to encompass non-periodic functions on $\mathbb{R}$, aiming to associate a function $f$ defined on $\mathbb{R}$ with another function $\hat{f}$ on $\mathbb{R}$. This $\hat{f}$ is referred to as the \textit{Fourier transform} \cite{gray2012fourier, kammler2007first, stein2011fourier} of $f$ and is defined by:

\begin{equation}
\hat{f}(\omega) = \int_{-\infty}^{\infty} f(x)e^{-2\pi ix\omega}dx
\end{equation}

Thus, extending a periodic function infinitely allows designating this transform $\hat{f}$ as the representation of $f$ in the frequency domain. One of the advantages of the Fourier transform is that is easily invertible, being in this case the equation \ref{eq:invFou}

\begin{equation}\label{eq:invFou}
   f(x) = \int_{-\infty}^\infty \hat{f}(\omega)e^{2\pi i n \omega} d\omega,
\end{equation}

In our context, we utilize the discrete version of the Fourier transform denoted by the equation:

\begin{equation}
X(n)=\sum_{m=0}^{N-1}x(m)e^{-\frac{2\pi in}{N}},
\end{equation}

where $N$ represents the total number of points. Given the objective of minimizing computational costs while building a robust aggregation function, the Fast Fourier transform (FFT) is employed. Instead of conducting a total of $N^2$ operations, this technique recursively divides the sum into even and odd points, effectively reducing the total operations to $N\log N$. From the comparison in \cite{shi2022challenges} illustrated in Table \ref{tab:comp}, it's evident that the FFT, our selected function, doesn't incur additional computation compared to others, highlighting the notably high computational cost associated with the Krumm function.


\begin{table}[]
\centering
\begin{tabular}{lc}
\textbf{}             & \textbf{Computation time}                          \\ \hline
\textbf{Mean}         & $\mathcal{O}(KN\log{N})$ \\\hline
\textbf{Median}       & $\mathcal{O}(KN\log{N})$ \\\hline
\textbf{Trimmed mean} & $\mathcal{O}(KN\log{N})$ \\\hline
\textbf{Krumm}        & $\mathcal{O}(KN^2)$      \\\hline
\textbf{FFT}          & $\mathcal{O}(KN\log{N})$\\\hline
\end{tabular}
\caption{Computation cost of the different aggregation functions}
\label{tab:comp}
\end{table}

\section{Related work}\label{related}
In recent years, the resilience of FL systems against poisoning attacks has garnered significant interest. Some recent works applied some of the functions described in the previous section to specific attacks using certain datasets. For example, \cite{rey2022federated} employs median and trimmed mean to protect their federated process using an intrusion detection dataset for IoT, although authors do not consider sophisticated colluding attacks, like the min-max. The description of median and trimmed mean is provided by \cite{yin2018byzantine}, but the authors do not offer experimental results on the resilience of these functions to Byzantine attacks. Additionally, \cite{so2020byzantine} implements a variant of Krum called multi-Krum, in which the update generated by the aggregator in each round is based on averaging several updates provided by the clients. While collusion attacks are considered, these are related to privacy aspects \cite{ruzafa2021intrusion}.

Moreover, other studies consider variations of existing aggregation functions or define new approaches to address different poisoning attacks. In this direction, \cite{pillutla2022robust} uses the geometric median instead of the median as employed in FedMedian. Specifically, the authors define RFA, a robust aggregation algorithm based on the weighted geometric median to defend the system against clients' corrupted weights considering aspects of privacy and efficiency. Moreover, the authors of \cite{li2021byzantine} extend this technique of the geometric median by incorporating additional weights for those that are close to each other based on their formula for measuring skewness. In both works, they adapted their method to fit non-IID scenarios by adding personalization techniques \cite{kulkarni2020survey}. Looking at other works that use different aggregation functions, in \cite{cao2020fltrust}, the authors present FLTrust, a method to prevent the global model from being infected by malicious clients. In their method, after each client iteration, the server checks for maliciousness by calculating their trust score based on similarity with the global model using cosine similarity, then they apply the ReLu function to take those greater than 0. Next, the weights are normalized and aggregated using the mean. Finally, the work \cite{li2021lomar} presents LoMar, an algorithm created to defend the FL process against poisoning attacks. Their algorithm is divided into two parts, first they divide the model weights using the $k$ nearest neighbors and then use their density using kernel density estimation (KDE) to calculate what they call the client's malicious factor. And in the second part, they set a threshold to assess which clients are malicious.

Unlike previous approaches, our proposed robust aggregation function utilizes the Fourier Transform to project the weights sent by clients in each training round into the frequency domain. Using the density function, values with the highest density are selected for aggregation. It is noteworthy that only recently a study by \cite{zhao2022transform} considered a transform-domain based approach to enhance efficiency and precision, though security aspects are not addressed. In our case, we use the fast Fourier Transform to convert the weights sent by the clients into the frequency domain. Our approach is tested against simple poisoning attacks in which a set of attackers generate random weights \cite{pillutla2022robust}, as well as in scenarios with sophisticated attacks where several compromised nodes collude using the min-max attack, which has been proven to be a highly effective attack approach in FL scenarios \cite{shejwalkar2021manipulating}.
\section{Threat model and considered attacks} \label{sec:Setup}
In this section, we describe the attacks considered in this work, and then, we will describe the complete FL system involving benign and malign clients.

\subsection{Threat model}
Based on the description provided in Section \ref{sec:Preliminaries}, we provide a detailed explanation of the considered FL scenarios, as well as the attacks implemented to assess our robust aggregation approach. Specifically, we consider an FL scenario composed of K clients and one aggregator. It is important to note that, although the proposed scenario only considers a single aggregator, our approach can be used in a decentralized or hierarchical setting where multiple aggregators are deployed \cite{lalitha2018fully}. Our threat model assumes the existence of a number M of compromised clients or byzantine attackers, where $M < K/2$, and we assume that the aggregator is not compromised. In each evaluation scenario, all malicious clients $m \in M$ will execute one of the two considered attacks, which are described in the following subsection.

To describe our threat model, we consider the systematization of threat models for poisoning attacks in FL proposed by \cite{shejwalkar2022back}. This system describes different aspects related to the characteristics of the attackers, including their objectives, knowledge, and capabilities:

\begin{itemize}
    \item \textbf{Objective of the adversary}: for both attacks, the goal is to reduce the model's accuracy by manipulating the weights generated from local training in each FL round. Therefore, our considered attacks are \textit{indiscriminate} (or \textit{untargeted} \cite{baruch2019little}) and \textit{generic}, as their aim is to cause a general misclassification of the data used, without targeting a specific class or set of samples.
    \item \textbf{Knowledge of the adversary}: in both attacks, the attackers only have access to the model updates produced by the client they control, but not to the other clients. Nevertheless, in the case of the min-max attack (see Section \ref{sec:attacks}), the compromised nodes have access to the updates produced by other malicious nodes to collude with them. Therefore, following the categorization described by \cite{shejwalkar2022back}, we assume \textit{whitebox} knowledge regarding the models produced by the clients in each training round. Moreover, although it could be considered in future work, we assume that the attackers do not have information related to the aggregation function used.
    \item \textbf{Capabilities of the adversary}: our considered attacks are based on poisoning the models produced by the clients in each training round. Access can be triggered by direct access to the device itself, or through attacks such as Man-in-the-middle where the attacker receives the updates and modifies them before they are sent to the aggregator in each training round. Furthermore, both attacks follow an \textit{online} mode, since, unlike typical attacks like label-flipping which are carried out at the beginning of training, our attacks are launched in every training round. In addition, in the case of the min-max attack, the attackers have the ability to collude with each other by exchanging their local updates to achieve a greater impact.
\end{itemize}

Additionally, we assume a scenario with no dropouts or \textit{stragglers} \cite{park2021sageflow}, meaning the nodes remain active throughout all the rounds of the training process.

\subsection{Attacks considered}\label{sec:attacks}
Before describing our robust aggregation approach for FL, we first outline the considered attacks. It is important to note that we consider that in each training round, a node may be compromised with the intention of degrading the performance of the global model. The first attack is based on sending random weights in each training round. Specifically, instead of sending updated weights based on local training on their data, a compromised node will send a set of weights generated randomly through a random Gaussian initialization (provided by the Keras library) to the aggregator. As demonstrated in previous works \cite{pillutla2022robust}, despite their low complexity, the implementation of these attacks leads to a significant degradation of performance if appropriate measures are not deployed.

The second type of attack is introduced in \cite{shejwalkar2021manipulating}, where the authors test some of the robust aggregation functions described in Section \ref{sec:Preliminaries}, and exploit their weaknesses to reduce their overall performance. 
The authors catalog various attack strategies, among which we adopt a particularly challenges variant: the agnostic \textit{min-max} attack. This strategy is designed to be effective even without knowledge of the aggregation function or the benign clients' weights. The choice to focus on the \textit{min-max} attack is that according to \cite{shejwalkar2021manipulating}, its impact is more significant compared with other attacks, such as LIE \cite{baruch2019little} and Fang \cite{fang2020local} attacks.

\begin{figure*} [!ht]
	\centering
		\includegraphics[width=2\columnwidth]{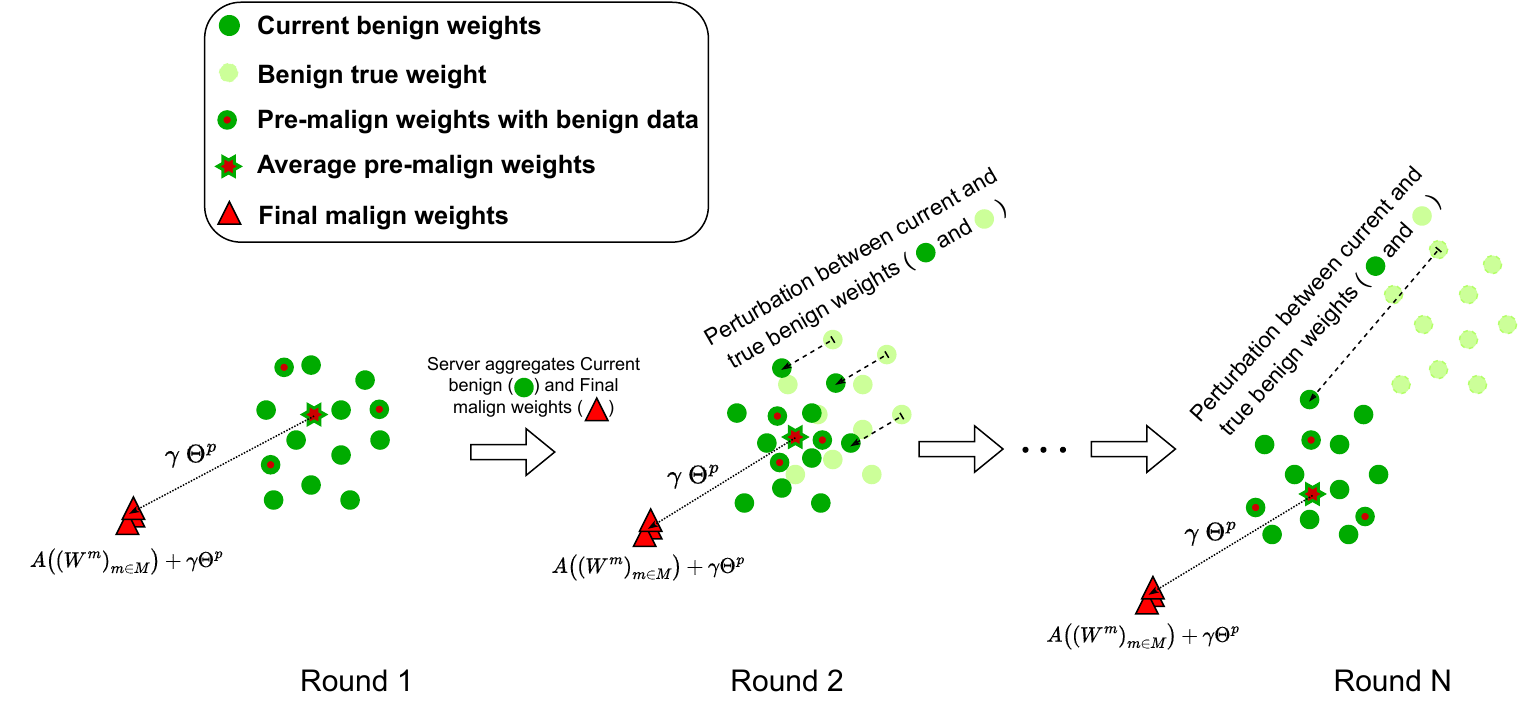}
	\caption{Visual description of how min-max attack works. The malicious weights (red points) separate from legitimate weights (green points), with the intention of degrading the overall system performance. The attack forces the legitimate weights to deviate from their true positions (indicated by light green), which they would naturally converge to in the absence of malicious entities.}
	\label{fig:minmax}
\end{figure*}

The methodology of min-max attack is based on an initial step in which malicious clients conduct model training with a standard dataset, like a benign client. Subsequently, these malign entities circulate their computed weights among themselves. The next step involves the calculation of new weights, which are crafted to maximize divergence from both the aggregated benign weights and from each other's weights, while maintaining a level of subtlety to eschew detection. These optimized weights are then sent to the aggregator. The equation to derive such final malign weights is described in Equation \eqref{eq:minmax}:
\begin{equation}
    \label{eq:minmax}
    \Theta= A((W^m)_{m\in M}) + \gamma\Theta^p,
\end{equation}

where $M$ denotes the set of malicious clients, $A(\cdot)$ is the aggregation function, $\Theta^p$ symbolizes the perturbation vector, and $\gamma$ is a scaling coefficient. Here, $\Theta^p$ and $\gamma$ represent the direction and the distance of the perturbation added to the average weights of malicious clients. $\gamma$ is calculated by \ref{eq:minmax2}. $\Theta^p$ can be calculated by three different ways \cite{shejwalkar2021manipulating}: through the inverse unit vector, i.e., dividing the weights through their norm, $ = -\frac{\Theta}{||\Theta||_2}$; the inverse standard deviation, $\Theta^p = -std((W^m)_{m\in M})$; and the inverse sign, $\Theta^p = -sign(\Theta)$, where $sign$ in equal to 1, 0, or -1 depending on the sign of the value. In this work, we use the inverse unit vector method, as it induced the most substantial decline in the global model's accuracy. A graphical representation of the min-max attack is provided in Figure \ref{fig:minmax}. Within this figure, individual points denote the weights: the malicious weights are in red, benign weights in solid green, weights of the malignant client post-training are green with a red dot, and light green points depict the hypothetical benign weights in the absence of any malign influence. These malign weights, as previously described, are strategically positioned as distant as feasible from the benign weights, incrementally causing divergence to reduce the global model's accuracy. To do so, after the training, the malign clients share their weights in order to calculate the mean, and then by equation \ref{eq:minmax} calculate the final malign weights to be sent to the server.

\begin{equation}
    \label{eq:minmax2}
    argmax_{\gamma} \max_{m\in M} ||\Theta -W^m|| \leq \max_{m,l \in M}  ||W^m - W^l||
\end{equation}

\begin{figure} [!ht]
	\centering
		\includegraphics[width=\columnwidth]{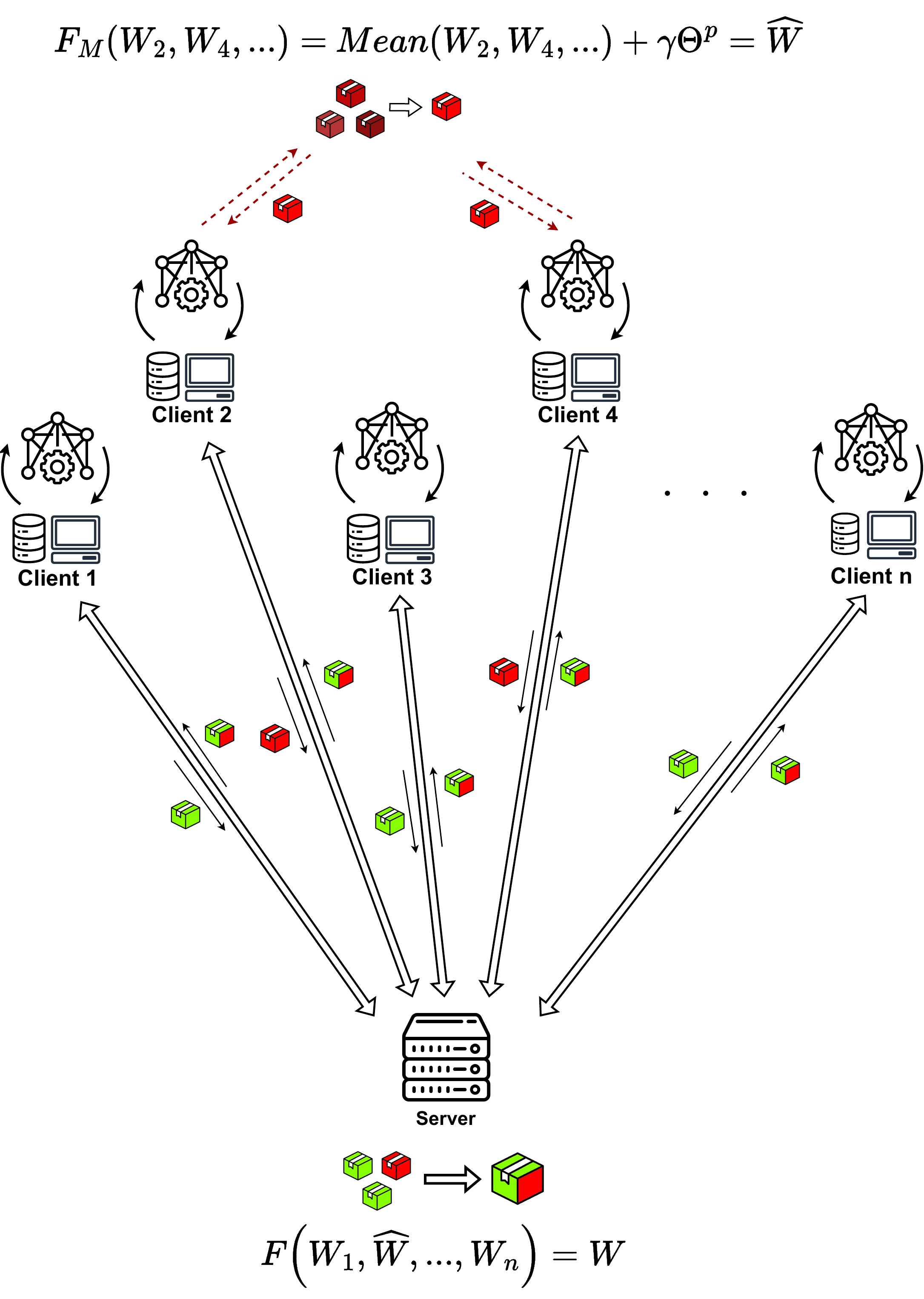}
	\caption{Scheme of our federated environment. Benign weights are presented as green packages. Malign clients cooperate to send malicious weights (red packages) to the server using a certain function $F_M$ (\ref{eq:minmax} in our case). The server aggregates the weights using a certain aggregation function $F$}
	\label{fig:FLsetup}
\end{figure}

The resulting FL scheme considered in this work is shown in Figure \ref{fig:FLsetup} with $K$ clients where there are $M<K$ compromised clients. These clients will send fake weights with the intention of harming the model performance, deviating the benign weights from their original direction. In some cases, malicious clients may operate independently (e.g., sending random weights) or collaborate to maximize their impact (e.g., through the min-max attack). Particularly in scenarios where several clients collude, as illustrated in Fig. \ref{fig:FLsetup}, certain clients join forces to generate certain weights aimed at significantly disrupting the overall process. This collaboration involves employing a specific function, denoted as $F_M$, which in our context refers to the min-max attack (see Equation \ref{eq:minmax}). 

\section{Methodology}\label{sec:methodology}
\begin{figure*} [!ht]
	\centering
		\includegraphics[scale = 0.95]{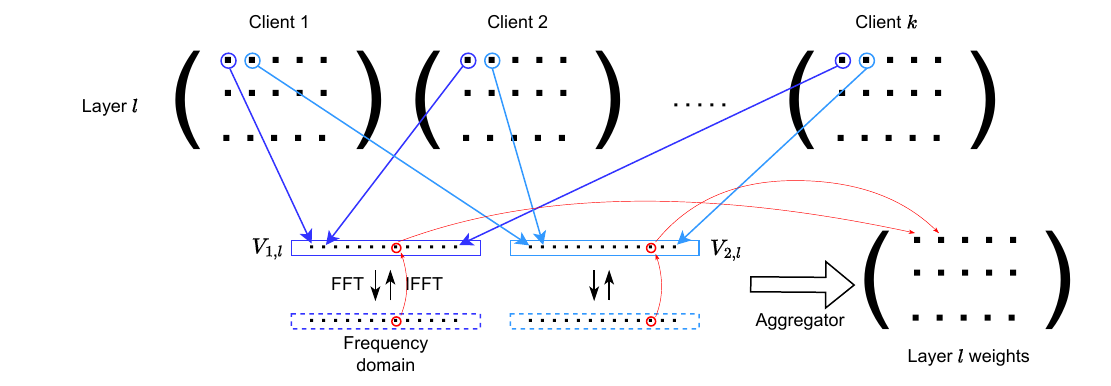}
	\caption{Visual description of our aggregation method}
	\label{fig:coorwise}
\end{figure*}

In this section, we describe our Byzantine-resilient secure aggregation function based on the FFT. An inherent property of the FFT is the capability of obtaining the density function through its frequency domain \cite{nanbu1995fourier}. From this density function, we select the maximum value, i.e., the point with highest frequency. For aggregating the clients' weights using the FFT, once the server receives the weights $(W^k)_1^K$ of the $K$ clients, it calculates the FFT in a coordinate-wise way  and selects the point whose FFT is the highest. Specifically, as shown in Fig. \ref{fig:coorwise} we calculate the vectors $V_{i,l} = \{w^1_{i,l}, \dots, w^K_{i,l} \}$, where $w^k_{i,l}$ is the $i$-th element of the weights $W^k$ at layer $l$ of client $k$. Then, these vectors $V_{i,l}$ are sorted, and for each $i$ and $l$, we compute the FFT of this vector to transform it into a frequency vector, since as said in section \ref{sec:fourier}, we can project any function to the frequency domain, and hence, calculate the density function in this case. Then, we take the index of the point of $V_{i,l}$ that achieves the maximum value in the frequency vector.

Specifically, we take:
\begin{equation}\label{eq:fourierfinal}
    \Bar{w}_{i,l} = arg\, max_{w^k_{i,l}}\{FFT(V_{i,l})\} \forall i,l
\end{equation}

Then, the aggregated model is calculated by joining all these $\Bar{w}_{i,l}$ into $\Bar{W}$. Such model $\Bar{W}$ is sent to the clients. This method operates on the principle that the weights from benign clients are likely to be grouped closely in the vector space. By applying the FFT to compute the density function of the vector $V_{i,l}$, we can identify such groups. The density function is expected to peak in regions where benign weights congregate, while malicious weights, which must deviate to impact performance adversely, will be isolated from these high-density zones. Consequently, such outlier weights exert minimal influence on the result of Equation \ref{eq:fourierfinal}, regardless of their magnitudes. 

This approach effectively shields the system against malicious clients. However, as discussed in Section \ref{related} and further in our work, in scenarios devoid of malicious activity, a simple mean of the weights proves to be the most efficient. Traditional methods prioritize defense against spurious weights at the expense of some accuracy, an unnecessary trade-off in the absence of malicious clients. To address this, our method incorporates a dynamic selection mechanism between the mean and Equation \ref{eq:fourierfinal}. This adaptability ensures minimal accuracy loss in any situation, whether malicious clients are present or not. Specifically, our approach uses the FedAvg function in environments without presence of malicious activity, shifting towards FFT-based aggregation as the presence of malicious clients increases.

To optimize the final accuracy of the model and judiciously select between the two aggregation methods, we employ the Kolmogorov–Smirnov (K-S) test \cite{berger2014kolmogorov}, as outlined in \ref{alg:pseudocode}. The K-S test is a robust, non-parametric method for assessing the goodness-of-fit. This test contrasts two sample distributions to determine if they significantly diverge from one another. The essence of the K-S test is captured by the following equation:
\begin{equation}
    KS(F_1,F_2) = sup{F_1(x) - F_2(x)}
\end{equation}

where $F1$ and $F2$ are the cumulative distribution functions of the two samples under comparison, and $sup$ denotes the maximum difference between these distributions. By applying this test, we can effectively discern between benign and potentially  malicious weight distributions, thereby guiding the decision to use either of the aggregation methods to improve the model's performance.

We operate under the assumption that benign weights, which are derived from using the same model and dataset (but different samples), will exhibit a consistent distribution pattern. To test this assumption, we select a subset of points of size $S$ from the vectors $V_{i,l}$ and use the remaining points to calculate the $p$-value from the K–S test. A $p$-value exceeding 0.05 indicates that the sampled subset adheres to the general distribution of the remaining points, suggesting the absence of malign influences. Conversely, a lower $p$-value implies a deviation from the expected distribution, signaling the presence of potentially malicious clients.

This testing procedure is replicated $C$ times for each vector $V_{i,l}$, with each iteration considering a different subset. We then compute the frequency with which the test results indicate a deviation from the expected distribution. The mean of these frequencies is calculated for every $i$ and $l$ of $V_{i,l}$. If this average $p$ surpasses a predetermined threshold $\alpha$, it implies the presence of malicious weights, prompting the server to implement the FFT for aggregation. If the average $p$ remains below $\alpha$, the server defaults to using FedAvg. A concise overview of this process is depicted in \ref{eq:trozos}, where $F$ represents the server's aggregation function.

\begin{equation}\label{eq:trozos}
    F(W^1,\dots,W^K)= \left\{ \begin{array}{lcc}
                FedAvg(W^1,\dots,W^K) & if & p \leq \alpha  \\ \\ 
                FFT(W^1,\dots,W^K)& if &p > \alpha  \end{array} \right.
\end{equation}

This methodology is designed not only to optimize performance irrespective of the number of malicious clients but also to safeguard the system against sudden attacks. For instance, consider a scenario where the system is secure at round $r$, but at round $r+1$, several clients become compromised and begin transmitting malicious weights. In such a situation, our approach would utilize FedAvg as aggregation method for all rounds up to and including round $r$. However, upon the onset of the attack at round $r+1$, it would switch to using the FFT for aggregation. This adaptive strategy ensures that our method aligns with the most effective function for each specific circumstance, thereby avoiding any unnecessary loss in accuracy.

\begin{algorithm}
\caption{Algorithm of our robust dynamic federated learning framework}
\label{alg:pseudocode}
\begin{algorithmic}[1]
\REQUIRE $N$ set of clients, $R$ number of rounds, $E$ number of epochs, $C$ number of repetitions of malicious test, $S$ size of the subset part of malicious test, and $t$ threshold.

\ENSURE Global model $W$ \\

\FOR{$r$ in 1 to $R$}
    \STATE $W_r^k = \text{localUpdate}(W_{r-1},E)$
    \STATE Send $W_r^k$ to the server

\vspace{5pt}
\FOR{client $\in N$} 
    \STATE $W_r^k = \text{localUpdate}(W_{r-1},E)$ 
    \STATE Send $W_r^k$ to the server
\ENDFOR

\vspace{5pt}
\FOR{Server}
    \STATE $t_r = \text{mean}(Mal\_test(\{W_r^1,\dots,W_r^{|N|}\}), C, S)$ 
    \IF{$t_r < t$}
        \STATE $W_r = \text{mean}(\{W_r^1,\dots,W_r^{|N|}\})$
    \ELSE
        \STATE $W_r = \text{FFT}(\{W_r^1,\dots,W_r^{|N|}\})$  
    \ENDIF
    \STATE Send $W_r$ to clients
\ENDFOR

\ENDFOR

\vspace{7pt}
$Mal\_test(\{W_r^1,\dots,W_r^{|N|}\}), C, S):$
\STATE Calculate $V_{i,l} = \{w^1_{i,l}, \dots, w^K_{i,l} \}\forall i,l$
\FORALL{$V_{i,l}$}
\FOR{1, 2, \dots, C}
\STATE Take sample $\hat{V}_{i,l}$ of $V_{i,l}$ of size S 
\STATE Calculate $p$-value of $KS(\hat{V}_{i,l}, V_{i,l}\setminus \hat{V}_{i,l})$
\ENDFOR \hfill \break
\STATE Calculate average of the previous $p$-values
\ENDFOR \hfill \break
\STATE Return vector of $p$-values

\end{algorithmic}
\end{algorithm}

\section{Evaluation}\label{sec:evaluation}
In this section, we describe the evaluation conducted to demonstrate the feasibility of our approach and its ability to fights against the described attacks. Thus, we provide a comprehensive evaluation comparing the performance of our previously described mechanism in section \ref{sec:methodology} with well-known techniques such as Krum and Trimmed mean, in scenarios where the described attacks are implemented.

\subsection{Settings and dataset}
For the evaluation of our robust dynamic aggregation function, we consider several federated scenarios running in a virtual machine with processor Intel(R) Xeon(R) Silver 4214R CPU @ 2.40GHz with 32 cores and 96 GB of RAM. Furthermore, our scenarios are deployed using the Flower implementation \cite{flower}.

Moreover, we use the FEMNIST dataset, which represents a federated version of EMNIST \cite{cohen2017emnist} created by LEAF \cite{caldas2018leaf} that is publicly available\footnote{\url{https://github.com/TalwalkarLab/leaf}}. FEMNIST is a dataset of handwritten characters used for image classification. It is the result of dividing the EMNIST dataset into 3550 non-iid clients. The data contains 62 classes (the numbers from 0 to 9, and 52 letters either in upper and lower case). 

We employ a Convolutional Neural Network (CNN) architecture composed by four hidden layers: three convolutional layers with kernel sizes of 8x8 with 8, 16, and 24 channels, followed by a fully connected layer consisting of 128 neurons. Each of these layers utilizes the ReLU activation function. The network culminates in an output layer of 62 neurons, employing a softmax activation function for multi-class classification.

For the experimental setup, we initially selected a cohort of 100 clients through a random sampling process. These clients were then fixed to ensure consistency across all experiments and procedural steps. Each client dataset encompasses 1315 samples across 47 classes, with an average entropy of 0.93. This entropy value, computed using the Shannon entropy formula \cite{bonachela2008entropy}, serves as an indicator of class distribution balance within each client's dataset. Additionally, we divided the dataset of each node into training data (80\%) and testing data (20\%).

To simulate adversarial scenarios, malicious clients were randomly chosen from within the pool of 100 clients. Importantly, the selection of these malicious clients varied with each experimental run to ensure robustness in the evaluation. This variation in the selection of malicious clients across runs allows for a comprehensive assessment of the model's resilience to different adversarial configurations.

\subsection{Comparison of robust aggregation functions}

In this section, our method is evaluated in comparison with the other aggregation functions discussed in Section \ref{sec:Preliminaries}. We focus on their performance against the two types of attacks outlined in Section \ref{sec:Setup}, varying the proportion of malicious clients from 0\% to 50\%. We use 50 training rounds for training the FL model. Our analysis extends to additional aspects, including the evolution of model accuracy over more rounds, and the impact of the dynamic method introduced in the preceding section, particularly in relation to threshold selection.

To facilitate a more direct comparison with other methods, we initially present results excluding our dynamic client detection method, that is, we focused solely on the FFT approach. With aggregation functions like Krumm and Trimmed Mean, which require specification of the number of malicious nodes, we adjust their settings to match the resistance level to the proportion of malicious clients present. Therefore, this could be considered as the \textit{best case} for those methods as we assume there is a previous step to calculate such number of malicious clientes. For instance, with 20\% malicious clients, we configure Krumm and Trimmed Mean to resist and exclude 20 clients, respectively. Furthermore, to enhance the reliability of our findings, each result in this section represents the average of five separate runs.

\subsubsection{Random weights attack}
\begin{figure} [!ht]
	\centering
		\includegraphics[width=\columnwidth]{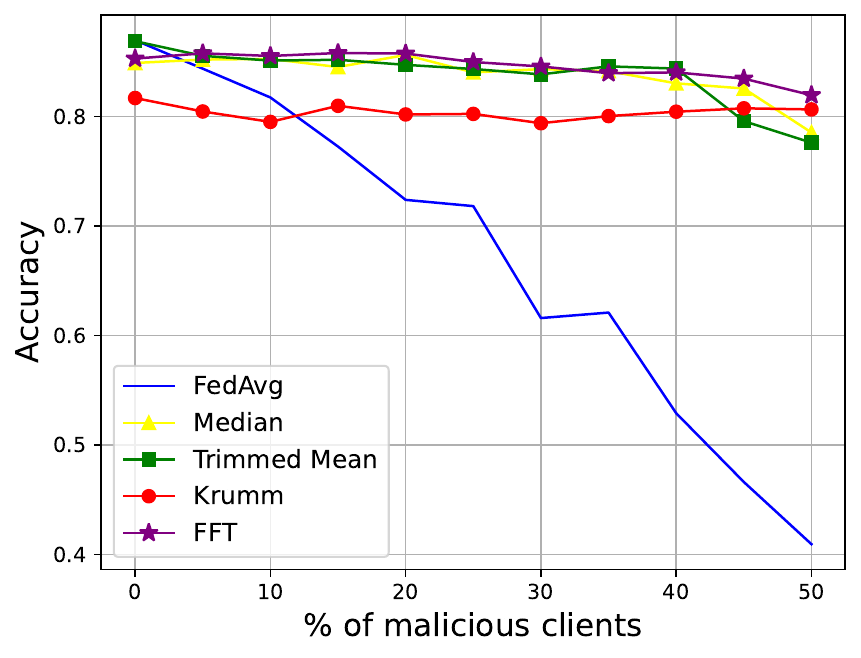}
	\caption{Comparison of accuracy different aggregation functions for random weights attack}
	\label{fig:comp_att_random}
\end{figure}

In evaluating the resilience of aggregation functions against an adversarial presence sending random weights, our work compares the accuracy of the different aggregation functions of section \ref{sec:Preliminaries} with the FFT. Table \ref{tab:attack_random} and Fig. \ref{fig:comp_att_random} show the performance obtained by the different aggregation functions. 

Under a scenario without malicious clients, FedAvg outperforms others with an accuracy of 0.8697. However, its efficacy declines rapidly with the introduction of adversarial actors. In the case of trimmed mean, it exhibits robustness up to a moderate level of malicious activity (around 40\%), post which it too succumbs to a substantial reduction in accuracy. Krum, while not suffering steep declines, consistently underperforms relative to the median and FFT across the spectrum of adversarial presence. This is particularly noteworthy as it suggests that Krum's mechanism for outlier detection does not translate into higher accuracy in the context of this specific type of attack.

Remarkably, both the median and FFT aggregation functions demonstrate comparable resistance against this attach, with FFT exhibiting slightly superior performance. This improvement is more pronounced beyond the 45\% threshold of malicious clients, where FFT sustains its accuracy more effectively than its counterparts. Therefore, our proposed FFT-based aggregation function manifests as the most robust across our tests. It maintains its accuracy above 0.80 even when malicious clients constitute up to 50\% of the network. These results underscore the potential of FFT as a robust aggregation strategy in FL environments with byzantine attackers.

\begin{table*}[]
\centering
\begin{tabular}{cccccc}
      \textbf{\% of malicious clients}            & \textbf{FedAvg}   & \textbf{Median} & \textbf{Trimmed mean} & \textbf{Krumm}  & \textbf{TFF}    \\ \hline
\textbf{0\%}      & \textbf{0,8697} & 0,8489          & 0,8688                & 0,8169          & 0,8529          \\ \hline
\textbf{5\%}      & 0,8437          & 0,852           & 0,8554                & 0,8046          & \textbf{0,8577} \\ \hline
\textbf{10\%}     & 0,8174          & 0,8526          & 0,8511                & 0,7951          & \textbf{0,8553} \\ \hline
\textbf{15\%} & 0,7726          & 0,8452          & 0,8518                & 0,8098          & \textbf{0,858}  \\ \hline
\textbf{20\%} & 0,7239          & 0,8562          & 0,8472                & 0,8019          & \textbf{0,8575} \\ \hline
\textbf{25\%} & 0,7182          & 0,8403          & 0,8434                & 0,8024          & \textbf{0,8498} \\ \hline
\textbf{30\%} & 0,616           & 0,8433          & 0,8385                & 0,7939          & \textbf{0,8457} \\ \hline
\textbf{35\%} & 0,621           & 0,8414 & \textbf{0,8458}                & 0,8004          & 0,8396          \\ \hline
\textbf{40\%} & 0,5291          & 0,8303          & 0,8438                & 0,8044          & \textbf{0,8403} \\ \hline
\textbf{45\%} & 0,4663          & 0,8256          & 0,7958                & 0,8075          & \textbf{0,8346} \\ \hline
\textbf{50\%} & 0,4095          & 0,7854          & 0,7761                & 0,8065          & \textbf{0,8195} \\ \hline
\end{tabular}
\caption{Accuracy of different aggregation functions against a random weights attack}
\label{tab:attack_random}
\end{table*}
			
\subsubsection{Min-max attack}
\begin{figure} [!ht]
	\centering
		\includegraphics[width=\columnwidth]{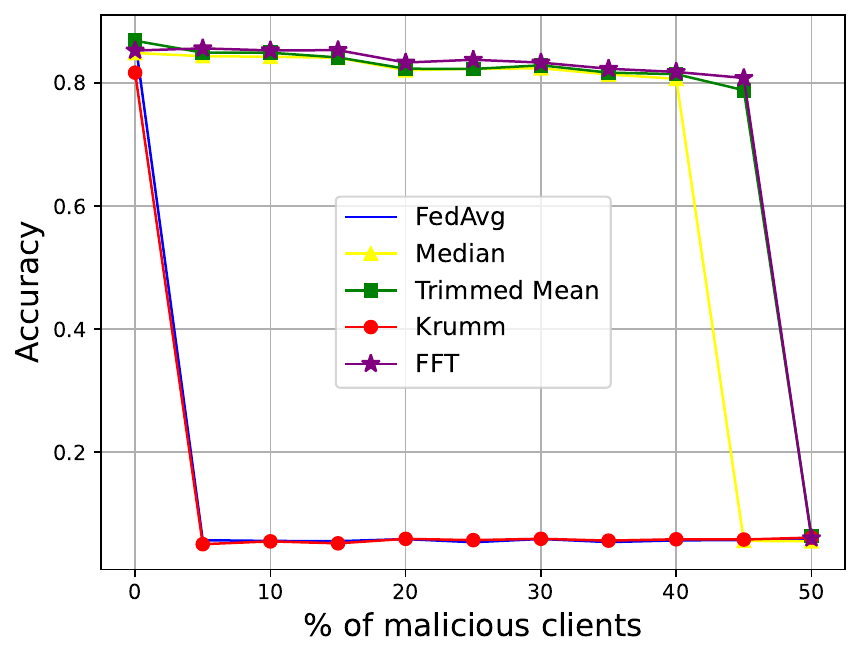}
	\caption{Comparison of different aggregation functions for min-max attack}
	\label{fig:comp_att_minmax}
\end{figure}

In the case of the more sophisticated min-max attack, 
the evaluation results are provided in Table \ref{tab:minmax} and Fig. \ref{fig:comp_att_minmax}. According to the results, FedAvg's performance deteriorates precipitously, even with a minimal 5\% incursion of malicious nodes, thereby highlighting its lack of robustness against this attack. Furthermore, the obtained results reveal that Krum's methodology of selecting weights based on minimizing the distance to its neighbors fails under this attack vector. Malicious clients exploiting this aspect can broadcast identical weights, resulting in a zero distance and consequently causing Krum to erroneously select these compromised weights. This flaw exposes a significant susceptibility of Krum to manipulation by colluding attacks

Regarding the median, trimmed mean, and our proposed FFT-based aggregation function, we observe a convergence in performance under moderate levels of malicious activity. However, the FFT approach provides better results by consistently achieving higher accuracy than the median and trimmed mean up to a threshold of 50\% of malicious clients. The resilience of FFT is noteworthy, declining from an accuracy of 0.8529 in a environment without malicious nodes to 0.8083 when nearly half of the clients are compromised.
Furthermore, it should be noted that trimmed mean's parameters were specifically tuned for each level of malicious activity, that is, the function assumes the exact number of malicious clients is known. This situation may not be feasible in real-world scenarios. This aspect accentuates the robustness of our FFT-based method, which does not rely on such tailoring to maintain superior performance against adversarial attacks.

\begin{table*}[]
\centering
\begin{tabular}{cccccc}
\textbf{\% of malicious clients} & \textbf{FedAvg} & \textbf{Median} & \textbf{Trimmed mean} & \textbf{Krumm} & \textbf{TFF} \\ \hline
\textbf{0\%}     & \textbf{0,8697} & 0,8489          & 0,8688                & 0,8169          & 0,8529       \\ \hline
\textbf{5\%}  & 0,0569          & 0,8437 & 0,8495          & 0,0504          & \textbf{0,8564} \\ \hline
\textbf{10\%} & 0,0557          & 0,8428 & 0,8493          & 0,0550          & \textbf{0,8531} \\ \hline
\textbf{15\%} & 0,0553          & 0,8411 & 0,8417          & 0,0518          & \textbf{0,8535} \\ \hline
\textbf{20\%} & 0,0587          & 0,8209 & 0,8234          & 0,0591          & \textbf{0,8334} \\ \hline
\textbf{25\%} & 0,0536          & 0,8230 & 0,8229          & 0,0570          & \textbf{0,8379} \\ \hline
\textbf{30\%} & 0,0589          & 0,8241 & 0,8289          & 0,0592          & \textbf{0,8333} \\ \hline
\textbf{35\%} & 0,0538          & 0,8140 & 0,8170          & 0,0562          & \textbf{0,8233} \\ \hline
\textbf{40\%} & 0,0566          & 0,8070 & 0,8145          & 0,0584          & \textbf{0,8181} \\ \hline
\textbf{45\%} & 0,0572          & 0,0558 & 0,7884          & 0,0580          & \textbf{0,8083} \\ \hline
\textbf{50\%} & 0,0592          & 0,0551 & \textbf{0,0637} & 0,0611          & 0,0601          \\ \hline
\end{tabular}
\caption{Accuracy of different aggregation functions against min-max attack}
\label{tab:minmax}
\end{table*}

\subsubsection{Evolution throughout training rounds}
\begin{figure*} [!ht]
	\centering
		\includegraphics[width=2\columnwidth]{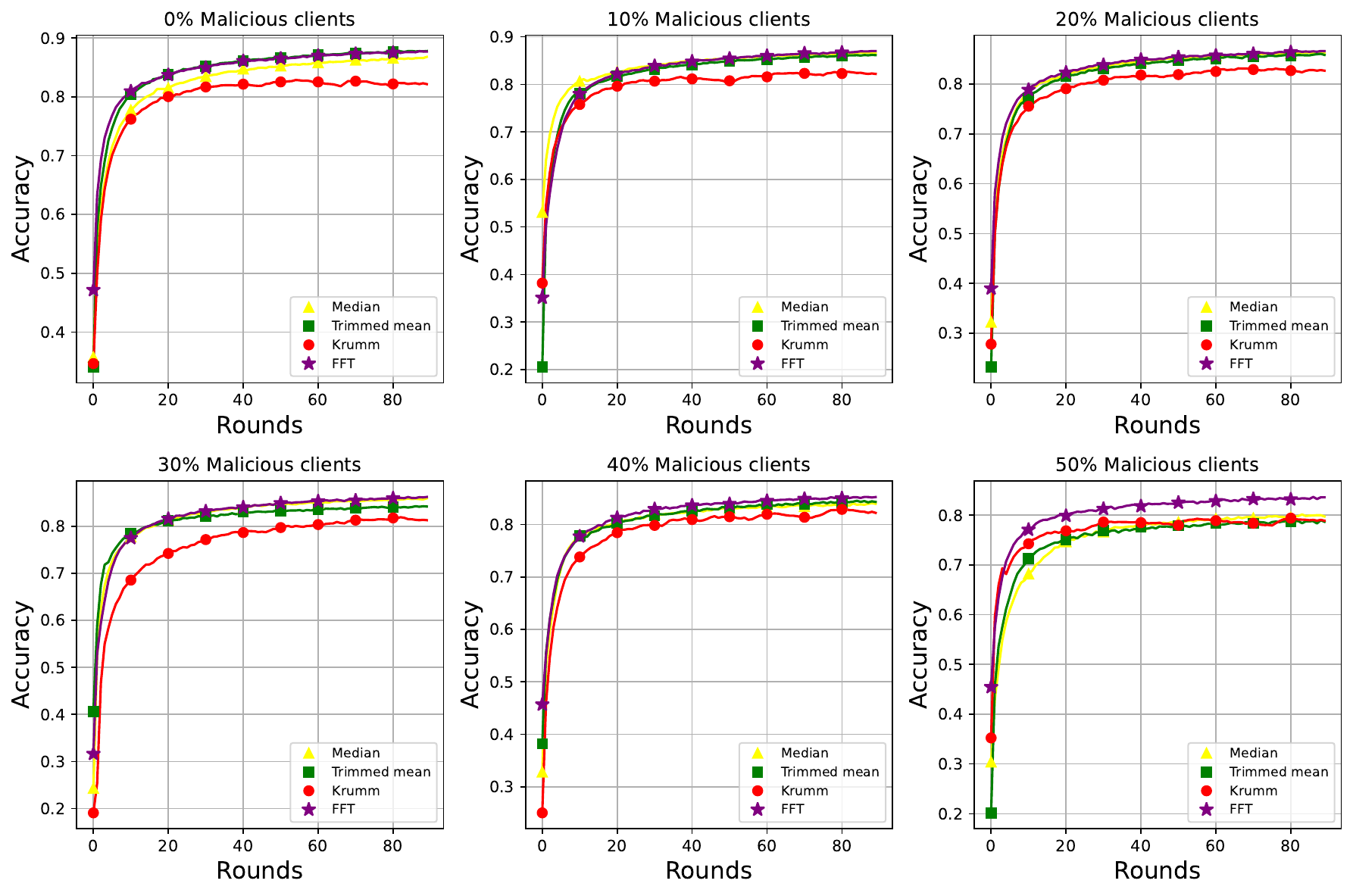}
	\caption{Analysis of evolution of the aggregation functions throughout the rounds for random weights attack}
	\label{fig:masRondas_at1}
\end{figure*}

\begin{figure*} [!ht]
	\centering
		\includegraphics[width=2\columnwidth]{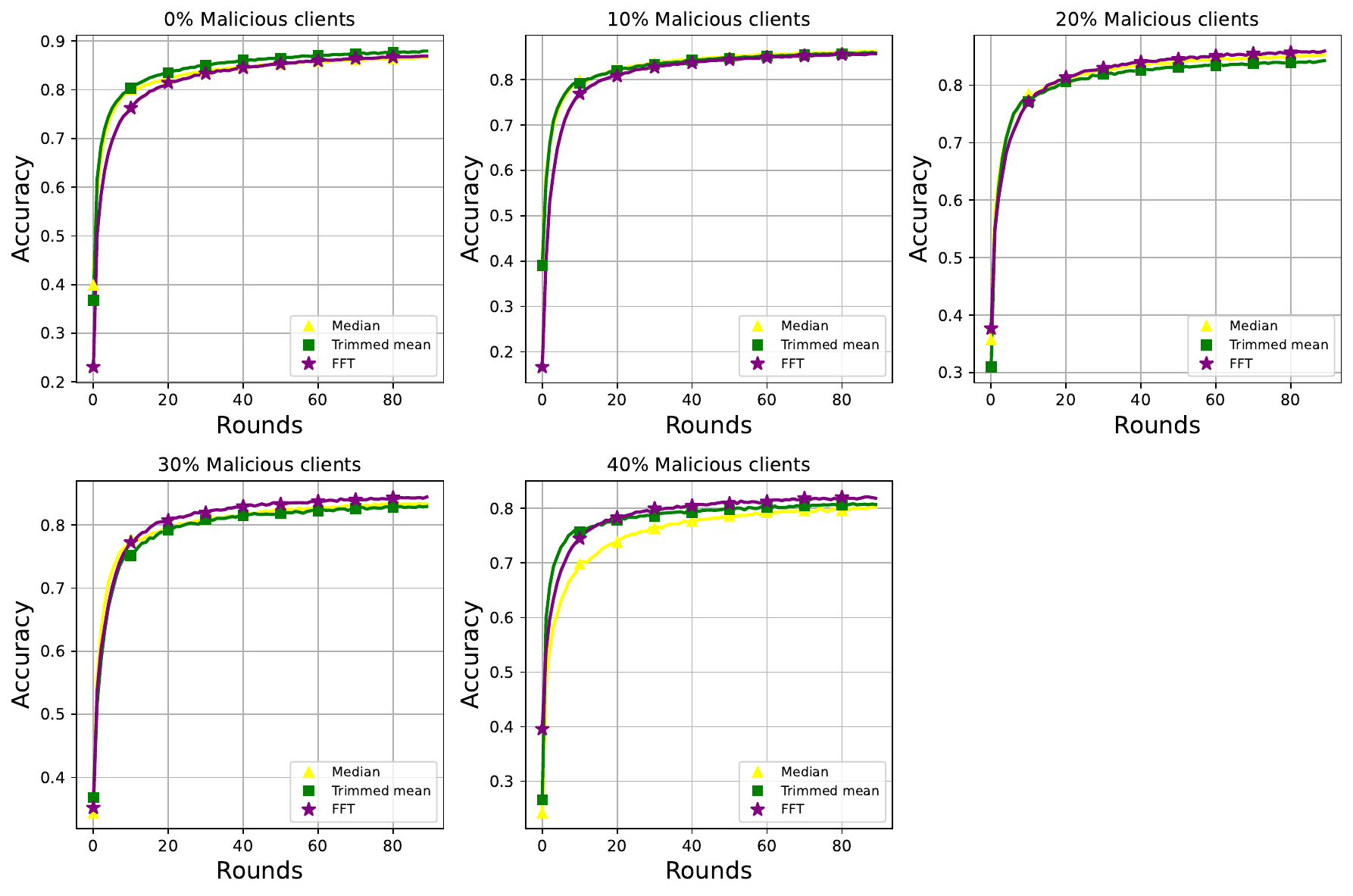}
	\caption{Analysis of evolution of the aggregation functions through the rounds for min-max attack}
	\label{fig:masRondas_at2}
\end{figure*}

Fig. \ref{fig:masRondas_at1} and \ref{fig:masRondas_at2} show the evolution of the performance of the different robust aggregation functions throughout the rounds against the two types of attack. The main goal is to show the convergence behavior of the different approaches under diverse attack scenarios. It should be noted that FedAvg is excluded from this analysis due to its rapid degradation in performance.

For clarity and to avoid overburdening the visual presentations, we modify the increments of malicious client representation to steps of 10\%, instead the previous 5\% increments. A total of 90 training rounds are considered to demonstrate the evolution of the different aggregation functions considering varying rates of malicious clients. For the random weights attack, the accuracy values of median, trimmed mean, Krumm, and FFT decrease from 0.8677, 0.8777, 0.8214, 0.8768 at 0\% of malicious clients to 0.7973, 0.7867, 0.7889, and 0.8358 at 50\% of malicious clients respectively. According to the results, for low rates of malicious clients, the different functions provide similar results except in the case of trimmed mean. 
The robustness of FFT becomes more pronounced as this rate increases. Indeed, with 50\% of malicious clients, the FFT-based approach truly stands out, maintaining higher accuracy than the other methods throughout the training rounds. This indicates that FFT is particularly effective in scenarios with a high level of adversarial presence. The median and Krum also perform well, significantly better than the trimmed mean, but not quite as effectively as FFT. This trend is also similar for the min-max attack when comparing FFT with median and trimmed mean. In particular, the accuracy values of median, trimmed mean, and FFT decrease from 0,8658, 0,8794, 0.8691 at 0\% of malicious clients to 0,8020, 0.8071, and 0.8185 at 50\% of malicious clients respectively. It should be noted that in this case, the Krumm approach is not shown due to the lack of resistance against this attack, as already mentioned. 

\subsubsection{Complete version of our algorithm}
As shown in Fig. \ref{fig:comp_att_random} and Fig. \ref{fig:comp_att_minmax}, when the proportion of malicious clients is at 0\%, FedAvg provides the highest value. However, its performance declines significantly as the presence of malicious clients increases. As already described in Section \ref{sec:methodology}, to maximize model accuracy regardless of the percentage of malicious clients, our approach involves dynamically toggling between FedAvg and FFT based on a threshold that estimates the level of malicious activity. This strategy is corroborated by the results presented in Tables \ref{tab:cv_at1} and \ref{tab:cv_at2}, which were derived using cross-validation with 5 folds across 4 threshold levels, as described by \cite{hastie2009elements}. This process ensures each sample is part of the test set at some point by systematically rotating the training and testing datasets.

\begin{figure} [!ht]
	\centering
		\includegraphics[width=\columnwidth]{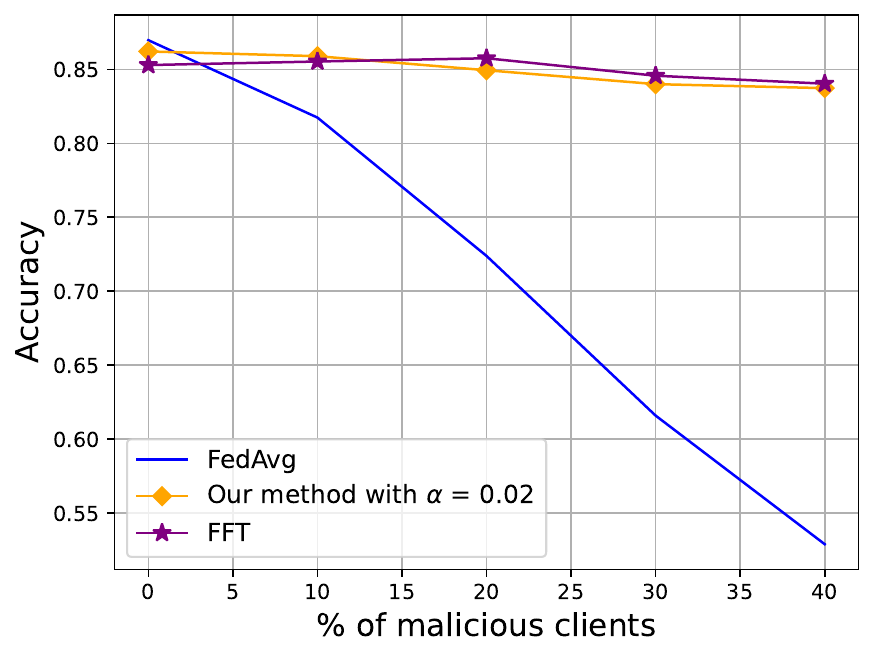}
	\caption{Comparison of our complete dynamic method with the static mean and FFT using the random attack}
	\label{fig:comp_cross_at1}
\end{figure}

\begin{figure} [!ht]
	\centering
		\includegraphics[width=\columnwidth]{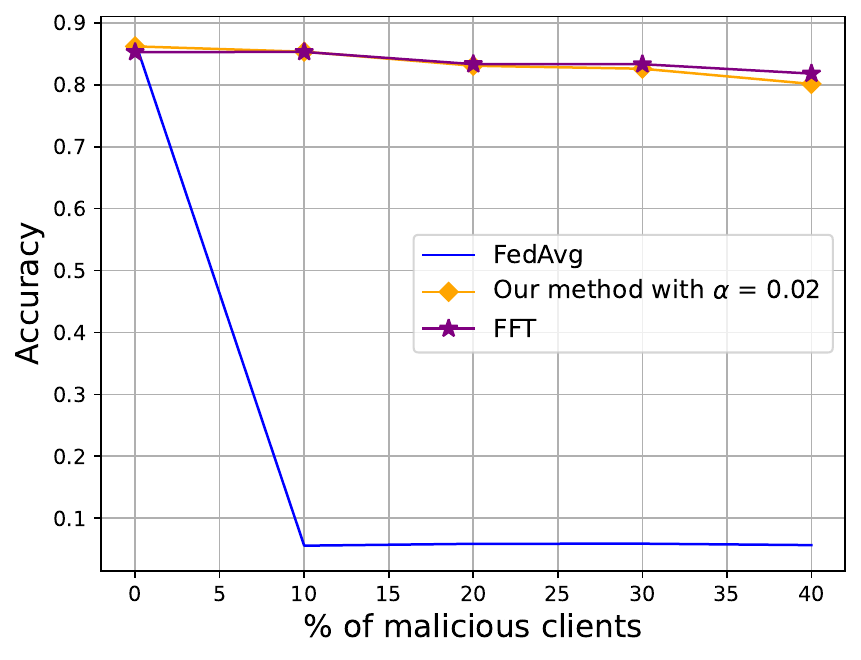}
	\caption{Comparison of our complete dynamic method with the static mean and FFT using the min-max attack}
	\label{fig:comp_cross_at2}
\end{figure}

Our findings indicate that a threshold value of $0.02$ yields the most favorable average outcome for both attack scenarios. Thus, in situations where the average skewness is less than or equal to $0.02$, we opt for FedAvg, and for higher skewness, the FFT-based approach is employed. In Figures \ref{fig:comp_cross_at1} and \ref{fig:comp_cross_at2}, we benchmark our dynamic threshold-based method, set at $0.02$, against the standalone performances of FFT and FedAvg from Figures \ref{fig:comp_att_random} and \ref{fig:comp_att_minmax}. These figures reveal that our method initially registers above FFT and slightly below to FedAvg at 0\% malicious clients, and subsequently falls below FFT as malicious clients emerge. Despite maintaining a minimal divergence from the optimal function, there is a noticeable gap at the last data point in both figures. This discrepancy arises because the K–S test, being probabilistic, may occasionally select an incorrect function, adversely affecting accuracy. Nevertheless, the overall efficacy of our dynamic method aligns with its intended purpose. It reliably discerns the presence of malicious clients in most instances, selecting the most suitable aggregation function for the given context. While the use of the K-S test can be seen as an initial approach for the identification of malicious clients, our future work is intended to come up with more sophisticated and precise approaches that could be further integrated with our proposed FFT-based aggregation function for a more robust overall FL framework.

\begin{table}[]
\centering
\begin{tabular}{ccccc}
              & \textbf{0,02}   & \textbf{0,025}  & \textbf{0,03} & \textbf{0,035}  \\ \hline
\textbf{0\%}  & 0,8621          & 0,8649          & 0,865         & \textbf{0,8678} \\ \hline
\textbf{10\%} & 0,8589          & \textbf{0,8598} & 0,8355        & 0,8374          \\ \hline
\textbf{20\%} & \textbf{0,8494} & 0,8289          & 0,7978        & 0,8058          \\ \hline
\textbf{30\%} & \textbf{0,84}   & 0,8088          & 0,7762        & 0,727           \\ \hline
\textbf{40\%} & \textbf{0,8373} & 0,7837          & 0,7441        & 0,6579          \\ \hline
\textbf{50\%} & \textbf{0,8004} & 0,7432          & 0,6112        & 0,5419          \\ \hline
\textbf{Mean} & \textbf{0,8413}          & 0,8148          & 0,7716        & 0,7396       \\  \hline 
\end{tabular}
\caption{cross validation for random attack}
\label{tab:cv_at1}
\end{table}

\begin{table}[]
\centering
\begin{tabular}{lllll}
            & \textbf{0,02}   & \textbf{0,025} & \textbf{0,03} & \textbf{0,035}  \\ \hline 
\textbf{0\%}  & 0,8621          & 0,8649         & 0,865         & \textbf{0,8678} \\\hline 
\textbf{10\%} & \textbf{0,8535} & 0,8476         & 0,8335        & 0,8108          \\\hline 
\textbf{20\%} & \textbf{0,8308} & 0,8066         & 0,7510        & 0,7077          \\\hline 
\textbf{30\%} & \textbf{0,8260} & 0,4933         & 0,6537        & 0,4005          \\\hline 
\textbf{40\%} & \textbf{0,8014} & 0,4580         & 0,2673        & 0,2696         \\\hline 
\textbf{Mean} & \textbf{0,8347} & 0,6940         & 0,6741        & 0,6112         \\\hline 

\end{tabular}
\caption{cross validation min-max attack}
\label{tab:cv_at2}
\end{table}

\section{Conclusions and future work}\label{sec:conclution}
This paper presented a novel approach to robust aggregation in FL, addressing the challenges associated with the capability to resist against model poisoning attacks. Unlike well-known techniques, such as median, trimmed mean, or Krum functions, our mechanism leverages the Fourier Transform to effectively filter and aggregate model updates in a FL scenario. Our method stands out by not requiring prior knowledge of the number of attackers and by being adaptable to dynamic and sophisticated attack scenarios. Through extensive evaluation against different types of model poisoning attacks, our approach has shown to outperform existing aggregation functions. The use of FedRDF not only enhances robustness against attacks but also maintains model accuracy in the absence of attackers, where FedAvg remains a preferable choice. Nevertheless, as part of our future work, we aim to explore more sophisticated approach beyond the K-S test to determine the existence of attackers and which specific clients are compromised to come up with a robust FL framework against different poisoning attacks.

\section*{Acknowledgements}
This study forms part of the ThinkInAzul program and was partially supported by MCIN with funding from European Union NextGenerationEU (PRTR-C17.I1) and by Comunidad Autónoma de la Región de Murcia - Fundación Séneca. It was partially funded by the HORIZON-MSCA-2021-SE-01-01 project Cloudstars (g.a. 101086248), the HORIZON-MSCA-2021-PF-01-01 project INCENTIVE (g.a. 101065524) and by the  ONOFRE Project PID2020-112675RB-C44 funded by MCIN/AEI/10.13039/501100011033 .

\small
\bibliographystyle{IEEEtranN}
\bibliography{biblio}

\begin{IEEEbiography}[{\includegraphics[width=1in,height=1.25in,clip,keepaspectratio]{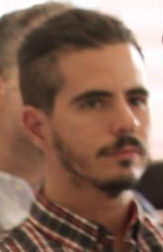}}]{Enrique Mármol Campos}
 is a Ph.D. Student at the university of Murcia. He graduated in Mathematics in 2018. Then, in 2019, he finished the M.S. in advanced math, in the specialty of operative research and statistic, at the university of Murcia. He is currently researching on federated learning applied to cybersecurity in IoT devices.
\end{IEEEbiography}

\begin{IEEEbiography}[{\includegraphics[width=1in,height=1.25in,clip,keepaspectratio]{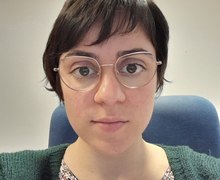}}]{Aurora Gonzalez Vidal}
 graduated in Mathematics from the University of Murcia in 2014. In 2015 she got a fellowship to work in the Statistical Division of the Research Support Service, where she specialized in Statistics and Data Analysis. Afterward, she studied a Big Data Master. In 2019, she got a Ph.D. in Computer Science. Currently, she is a postdoctoral researcher at the University of Murcia. She has collaborated in several national and European projects such as ENTROPY, IoTCrawler, and DEMETER. Her research covers machine learning in IoT-based environments, missing values imputation, and time-series segmentation. She is the president of the R Users Association UMUR.
\end{IEEEbiography}

\begin{IEEEbiography}[{\includegraphics[width=1in,height=1.25in,clip,keepaspectratio]{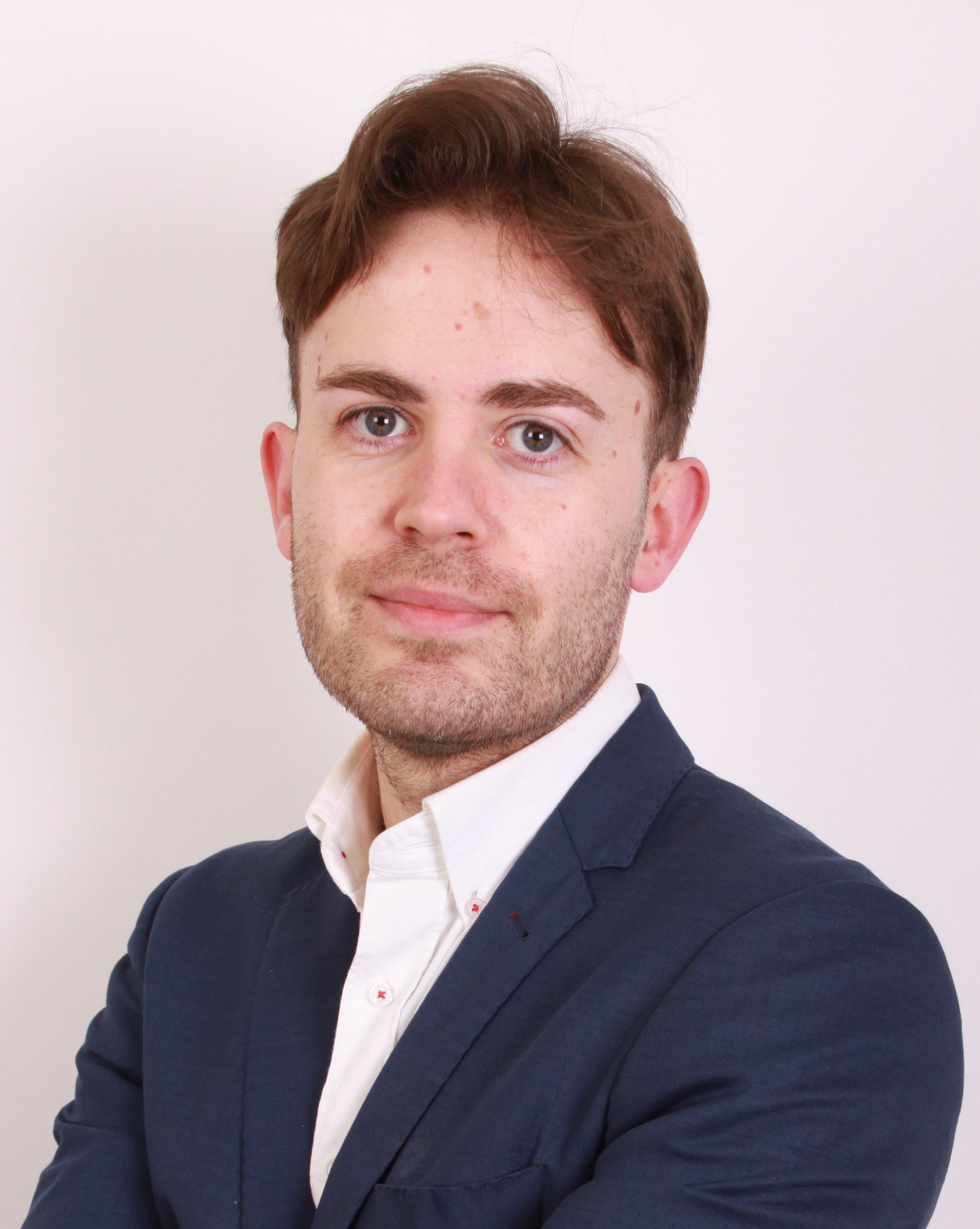}}]{José Luis Hernández Ramos}
 received the Ph.D. degree in computer science from the University of Murcia, Spain. He is a Marie Sklodowska-Curie Postdoctoral Fellow at the same university. Before that, he worked as a Scientific Project Officer at the European Commission. He has participated in different European research projects, such as SocIoTal, SMARTIE, and SerIoT, and published more than 60 peer-reviewed papers. His research interests include application of security and privacy mechanisms in the Internet of Things and transport systems scenarios, including blockchain and machine learning. He has served as a technical program committee and chair member for several international conferences, and editor in different journals.
\end{IEEEbiography}

\begin{IEEEbiography}[{\includegraphics[width=1in,height=1.25in,clip,keepaspectratio]{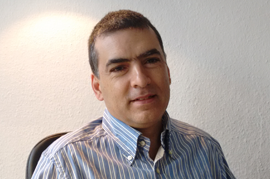}}]{Antonio Skarmeta}
 is a full professor at the University of Murcia, the Department of Information and Communications Engineering. His research interests are the integration of security services, identity, the Internet of Things, and smart cities. Skarmeta received a Ph.D. in computer science from the University of Murcia. He has published more than 200 international papers and been a member of several program committees.
\end{IEEEbiography}

\end{document}